\documentclass[runningheads,a4paper]{llncs}

\usepackage{amssymb}
\usepackage{graphicx}

\usepackage{epstopdf}
\usepackage{subfigure}
\usepackage{amsmath}
\usepackage{algorithm}
\usepackage{algorithmic}
\usepackage[algo2e]{algorithm2e}

\usepackage[table]{xcolor}
\definecolor{mydarkred}{rgb}{0.6,0,0}
\definecolor{mydarkgreen}{rgb}{0,0.6,0}
\usepackage[colorlinks,
linkcolor=mydarkred,
citecolor=mydarkgreen]{hyperref}
\usepackage{url}

\DeclareMathOperator{\tr}{tr}
\DeclareMathOperator{\nullspace}{nullspace}

\usepackage{url}
\urldef{\mailsa}\path|bo.han@riken.jp|

\begin{document}

\mainmatter  

\title{DATELINE: Deep Plackett-Luce Model with Uncertainty Measurements}

\titlerunning{Deep Plackett-Luce Model with Uncertainty Measurements}

%
%
\author{Bo Han$^{1,2}$\thanks{Preprint. Work in progress.}%
}
\authorrunning{Bo Han}

\institute{
$^1$Center for Advanced Intelligence Project, RIKEN, Japan\\
$^2$Centre for Artificial Intelligence, University of Technology Sydney, Australia\\
}

%
%
\maketitle

\begin{abstract}
The aggregation of $k$-ary preferences is a historical and important problem, since it has many real-world applications, such as peer grading, presidential elections and restaurant ranking. Meanwhile, variants of Plackett-Luce model has been applied to aggregate $k$-ary preferences. However, there are two urgent issues still existing in the current variants. First, most of them ignore feature information. Namely, they consider $k$-ary preferences instead of instance-dependent $k$-ary preferences. Second, these variants barely consider the uncertainty in $k$-ary preferences provided by agnostic crowds. In this paper, we propose \textbf{D}eep pl\textbf{A}cke\textbf{T}t-luce mod\textbf{EL} w\textbf{I}th u\textbf{N}certainty m\textbf{E}asurements (DATELINE), which can address both issues simultaneously. To address the first issue, we employ deep neural networks mapping each instance into its ranking score in Plackett-Luce model. Then, we present a weighted Plackett-Luce model to solve the second issue, where the weight is a dynamic uncertainty vector measuring the worker quality. More importantly, we provide theoretical guarantees for DATELINE to justify its robustness.
\end{abstract}

\section{Introduction}
The aggregation of $k$-ary preferences is a historical problem \cite{barbera1978preference}, and still keeps vibrant in recent years \cite{bottero2018choquet,li2018hybrid}. Besides, the aggregation of $k$-ary preferences has many real-world applications, such as peer grading \cite{raman2014methods}, presidential elections \cite{bartels1996uninformed} and restaurant ranking \cite{dwork2001rank}. Mathematically, score-based models can be leveraged to aggregate multiple $k$-ary preferences effectively \cite{volkovs2012flexible}.

For example, variants of Bradley-Terry model can indirectly aggregate $k$-ary preferences \cite{chen2013pairwise}, when $k$-ary preferences have been split into multiple pairwise preferences by the rank-breaking strategy \cite{khetan2016data,soufiani2014computing}. Nonetheless, inappropriate rank-breaking strategy will lead to inconsistent estimates \cite{han2018robust}. This issue motivates us to use variants of Plackett-Luce model \cite{guiver2009bayesian,maystre2015fast}, which can directly aggregate $k$-ary preferences.

However, there are two urgent issues still existing in the current variants of Plackett-Luce model. First, most of them ignore feature information. Namely, they only consider the order of preferences (a.k.a, object comparison), instead of considering the instance information corresponding to preferences. For instance, when aggregating $k$-ary preferences of face microexpressions \cite{yan2013fast}, traditional Plackett-Luce models fail to consider high-dimensional features of face.

Second, these variants barely consider the uncertainty in $k$-ary preferences provided by agnostic crowds. Specifically, $k$-ary preferences usually come from multiple people instead of sole one. When multiple people involve in such ranking procedure, they may introduce the uncertainty in $k$-ary preferences. Namely, $k$-ary preferences provided by agnostic crowds tend to become noisy, which will degrade the generalization of traditional Plackett-Luce models \cite{cheng2010label}.

In this paper, we propose \textbf{D}eep pl\textbf{A}cke\textbf{T}t-luce mod\textbf{EL} w\textbf{I}th u\textbf{N}certainty m\textbf{E}asurements (DATELINE), which can address both issues simultaneously. To address the first issue, we employ deep neural networks mapping each instance into its corresponding ranking score in Plackett-Luce model. Our target is to derive a more accurate aggregation model based on both object comparison and feature information. Furthermore, we present a weighted Plackett-Luce model to solve the second issue, where the weight is a dynamic uncertainty vector measuring the worker quality. The weight can be iteratively updated by feeding k-ary noisy preferences. Our target is to derive a more robust aggregation model based on the worker quality. In addition, we provide theoretical guarantees (i.e., minimax rates) for DATELINE to justify its robustness.

The remainder of this paper is organized as follows. Section~\ref{newsetting} provides a new $k$-ary preferences setting, namely \textit{instance-dependent noisy} preferences. Section~\ref{PLmodel} revisits the mediocre Plackett-Luce model from a stagewise perspective, and discloses its intrinsic deficiencies for handling instance-dependent noisy preferences setting. Section~\ref{DATELINE} proposes our core model DATELINE. Section~\ref{theory} provides theoretical guarantees related to DATELINE. Section~\ref{conclusion} concludes the current progress and discusses future works.

\begin{table}[!tp]
     \renewcommand{\arraystretch}{1.0}
     \normalsize
     \scalebox{0.9}{
     \begin{tabular}{ c l c }
     \hline
          Notation & Explanation\\ \hline
          $\Omega$                  &      set of all objects, $\Omega = \{O_1,O_2,\cdots,O_L\}$\\
          $\mathbf{x}_i \in \mathbb{R}^d$      &      the $d$-dimensional features of object $O_i$\\
          $\xi$                     &      subset of $\Omega$, $\xi\subseteq  \Omega$\\
          $L$                       &      $|\Omega|$, total number of all objects\\
          $W$                       &      number of crowd workers\\
          $D$                       &      collection of all $k$-ary preferences\\
          $D_w$                     &      collection of $k$-ary preferences annotated by crowd worker $w$\\
          $N_w$                     &      $|D_w|$, number of $k$-ary preferences annotated by crowd worker $w$\\
          $\rho_{n,w}$              &      the $n^{th}$ $k$-ary preference annotated by crowd worker $w$\\
          $l_{\rho_{n,w}}$          &      the length of preference $\rho_{n,w}$\\
         $\max(\xi)$                &      the best object in subset $\xi$ according to a criterion\\
          $O_i >O_j$                &      the ground truth order between $O_i$ and $O_j$\\
    $O_i \ \tilde{ > }\   O_j$      &      the preference annotated by crowd worker\\
    $\theta$      &   instances-shared parameter\\
    $\lambda_{i,\theta}$      &   instance-dependent ranking score for object $i$\\\hline
     \end{tabular}}
     \vskip 0.1in
     \caption{Common notations.}
\label{notation}
\end{table}

\section{New setting: Instance-dependent noisy preferences}\label{newsetting}

\begin{figure}[!tp]
\begin{center}
\centerline{\includegraphics[width=1\textwidth]{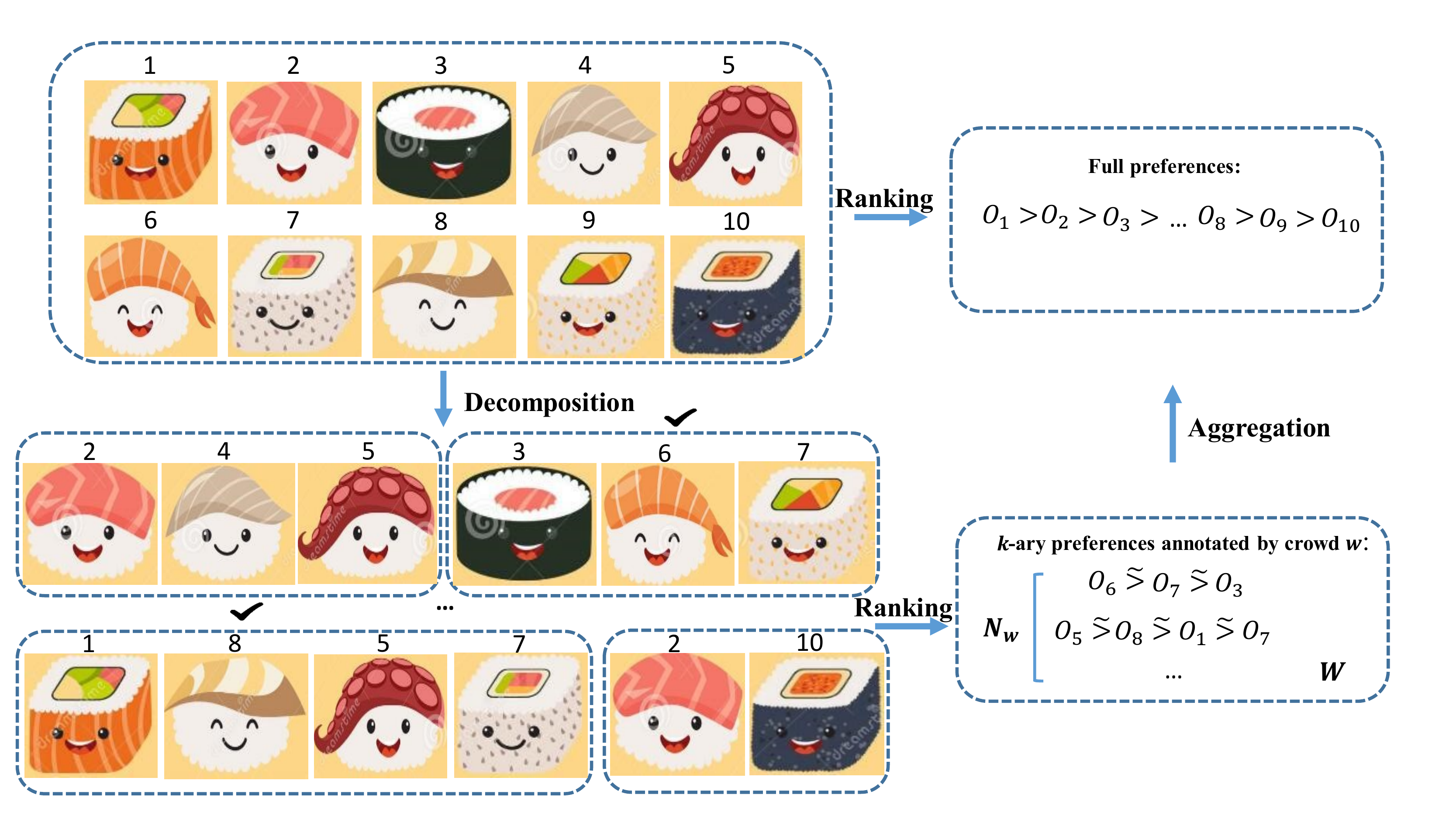}}
\caption{Instance-dependent noisy preferences setting (i.e., sushi ranking). Decomposition: a large set of objects is split into several subsets; Ranking: by considering feature information of each object, workers rank multiple (overlapped) subsets independently to yield $k$-ary preferences; Aggregation: aggregation methods aggregate multiple instance-dependent noisy preferences into a global preference. Note that: (1)~Feature information of each object affects the annotation process. (2)~The tasks (subsets) with ``{\large\checkmark}'' are assigned to the worker $w$. (3)~The notation $W$ in the corner denotes that $W$ workers complete the annotation process independently.}
\label{data-structure}
\end{center}
\end{figure}

Before delving into our new setting, we state and illustrate common notations in Table~\ref{notation}. Traditional preferences aggregation has two obvious characteristics: 1) Each worker disregards object features, and only focuses on object comparisons; 2) Each worker ranks his/her most confident $l$ objects and leaves the remaining $L-l$ objects undefined. Therefore, to reach a more accurate preference aggregation, we propose a new but practical setting called ``instance-dependent noisy preferences'' (Figure~\ref{data-structure}), which should meet two requirements as follows.

\begin{itemize}
  \item Features of each object should be considered, because they affect object comparisons, especially for image and natural language comparisons. For example, to compare the quality of three assignments, the position of each assignment should be decided by the text contents and the subjective bias simultaneously.
  \item Each worker only annotates multiple $k$-ary preferences, where the size of $k$ is not only dynamic, but also $k \ll L$. Specifically, dynamic $k$ is more flexible in the data collection. Meanwhile, $k \ll L$ makes sure that the worker has the sufficient confidence to finish each annotation.
\end{itemize}

\section{Plackett-Luce model}\label{PLmodel}

\subsection{A stagewise perspective}
Here, we revisit the Plackett-Luce model from the stagewise perspective~\cite{volkovs2012flexible}, which constructs a preference by a series of sequential stages. In each stage, compared to all the remaining alternatives, the object selected preferentially (without replacement) is regarded as the ``local winner''.

Following the stagewise learning strategy, Plackett-Luce model decomposes each $k$-ary preference into a series of sequential stages, and models each stage independently. Therefore, the likelihood function for the $k$-ary preference $\rho$ can also be expressed as follows:
\begin{equation}\label{stage-wise}
P(\rho | \vec{\lambda}) =  \prod_{i=1}^{k}P\left(X = \rho^{(i)} | \vec{\lambda}\right) = \prod_{i=1}^{k}\delta (\lambda_{\rho^{(i)}}),
\end{equation}
where $X \overset{\Delta}{=} \max (\rho^{(i)},\rho^{(i+1)},  \cdots, \rho^{(k)})$, indicating the local winner at stage~$i$. Furthermore, we use the normalized function $\delta (\lambda_{\rho^{(i)}}) =  \frac{\lambda_{\rho^{(i)}}}{\sum_{t=i}^{k} \lambda_{\rho^{(t)}}} $ to  model the probability \cite{tkachenko2016plackett} that object $\rho^{(i)}$ is selected as the local winner at stage~$i$.

\begin{remark}
For a preference $\rho$ annotated by crowd worker $w$, object $\rho^{(i)}$ is more preferable by worker $w$ than object $\rho^{(j)}$ $\forall i < j$.
\end{remark}

\subsection{Deficiency of Plackett-Luce model}

However, to handle the new setting proposed in Section~\ref{newsetting}, the direct usage of Plackett-Luce model has some essential deficiencies as follows.

\begin{itemize}
  \item The current model disregards objective features, and only focus on objective comparisons. This is unreasonable and should be corrected, especially for high-dimensional datasets (i.e., vision and language).
  \item The current model regards each $k$-ary preferences equally, which is unsuitable. Expert workers have a clear understanding about the contrast among objects, and they can make a confident decision when they annotate the preferences. However, amateur workers may annotate the preferences erroneously, due to their limited expertise about the contrast among objects.
\end{itemize}

\section{Deep Plackett-Luce model with uncertainty measurements}\label{DATELINE}

\subsection{Instance-dependent scores}

One limitation of the Plackett-Luce model is that this model depends on the object-specific parameters $\lambda_{i}$. However, for many tasks (i.e., image ranking, text ranking, and video ranking), we hope that the model is related to instances-shared parameter $\theta$ and high-dimensional instance $\mathbf{x}_i$ jointly. Namely, the ranking score $\lambda_{i,\theta}$ is instance-dependent as follows.

\begin{equation}\label{instance-scores}
\lambda_{i,\theta} = \exp(f_{\theta}(\mathbf{x}_i)),\: i \in \{1, \cdots, L\},
\end{equation}

where $f_{\theta}(\cdot)$ is a non-convex deep neural networks parameterized by $\theta$. Therefore, we bring the ranking parameter $\lambda_{i,\theta}$ into the feature space of objects $\mathbf{x}_i$.

\subsection{Uncertainty measurements}

Due to crowd workers' hesitation in selecting the local winner at each stage, stagewise learning strategy yields some deviations in modelling the noisy preferences.

To capture crowd workers' vacillation at each stage, we no longer exclusively rely on the single local winner selected by crowd workers, but consider other potential candidates of the local winner. To model the worker quality, we introduce an uncertainty vector $\vec{\eta}_w$ for each crowd worker $w$. The length of $\vec{\eta}_w$ for any crowd worker $w$ is set to the maximal preference length $K$, where $K = \max_{n,w}l_{\rho_{n,w}}$, $w = 1,2,\cdots,W$ and $n = 1,2,\cdots,N_w$.

Furthermore, we assume $\vec{\eta}_w = [\eta^{1}_w,\eta^{2}_w,,\cdots,\eta^{K}_w]$ with $\sum_{t=1}^K \eta^t_w =1$ for each crowd worker $w$, where entry $\eta^t_w$ represents the conditional probability that he/she selects the $1^{st}$-ranked object as the local winner given the real ground truth ranked at $t^{th}$. Our robust stagewise learning strategy avoids the deficiency of permutation-based approach, which significantly reduces the parameter space from $K!$ to $K$ accordingly.

However, for a $k$-ary preference $\rho:O_1 \ \tilde{ > }\   O_2 \ \tilde{ > }\   \cdots \ \tilde{ > }\  O_{k}$, there are different number of objects to compare at different stages, which causes different entries of the uncertainty vector being active at each stage. Therefore, a single uncertainty vector is not suitable for all stages simultaneously. To avoid this issue, we normalize the active entries at each stage, and popularize the definition of uncertainty vector to more general situations accordingly.

For the general case of stage~$i$, we have $(k-i+1)$ candidates, less than the maximal preference length $K$. Only the top $(k-i+1)$ entries of $\vec\eta_w$ are active. Then, we apply the renormalization trick on the active entries $[\eta^{1}_w,\eta^{2}_w,\cdots,\eta^{(k-i+1)}_w]$,  and generalize the definition of uncertainty vector accordingly.

\begin{remark}
We have the following observations: (1) For an expert worker $w$, $\eta^{t}_w$ decreases exponentially with $t$, as he/she has a clear understanding about the contrast among the objects. (2) For an amateur worker $w$, he/she may hesitate over comparable objects due to limited expertise. Namely, $\eta^1_w$, denoting the conditional probability that the selected local winner accords with the ground truth, does not gain the absolute advantage over other entries~$\eta^t_w (t \geq 2)$, especially~$\eta^2_w$.
\end{remark}

\subsection{DATELINE model}

After integrating the instant-dependent Plackett-Luce model with the introduced uncertainty vector, the likelihood of the $k$-ary preference $\rho$ at stage $i$ can be represented as:
\begin{equation}\label{uncertainty-vector}
\begin{split}
P\left(\widetilde{X} = \rho^{(i)}|\vec{\lambda}_{\theta},\vec{\eta}_w\right) & = \sum_{t=i}^{k}P\left(\widetilde{X} = \rho^{(i)}| X = \rho^{(t)}\right)P\left(X = \rho^{(t)} | \vec{\lambda}_{\theta}\right)\\
& = \sum_{t=i}^{k}\bar{\eta}^{(t-i+1)}_w\delta (\lambda_{\rho^{(t)},\theta}).
\end{split}
\end{equation}

Combining Eq.~\eqref{stage-wise}, Eq.~\eqref{instance-scores} and Eq.~\eqref{uncertainty-vector}, we propose our \textbf{D}eep pl\textbf{A}cke\textbf{T}t-luce mod\textbf{EL} w\textbf{I}th u\textbf{N}certainty m\textbf{E}asurements (DATELINE) for a collection of instance-dependent noisy preferences $D$, which can be expressed as follows:
\begin{equation}\label{couple}
\begin{split}
P(D|\vec{\lambda}_{\theta},\{\vec{\eta}_w\}_{w=1}^W) &=  \prod_{w=1}^W P(D_w|\vec{\lambda}_{\theta},\vec{\eta}_w)= \prod_{w=1}^W \prod_{n=1}^{N_w}P(\rho_{n,w} | \vec{\lambda}_{\theta},\vec{\eta}_w) \\
&= \prod_{w=1}^W \prod_{n=1}^{N_w}\prod_{i=1}^{l_{\rho^{}_{n,w}}} P\left(\widetilde{X} = \rho_{n,w}^{(i)}|\vec{\lambda}_{\theta},\vec{\eta}_w\right)\\
&= \prod_{w=1}^W \prod_{n=1}^{N_w}\prod_{i=1}^{l_{\rho^{}_{n,w}}} \sum_{t=i}^{l_{\rho^{}_{n,w}}}\bar{\eta}^{(t-i+1)}_w\delta (\lambda_{\rho_{n,w}^{(t)},\theta}),
\end{split}
\end{equation}
where $\vec{\eta}_w$ is the uncertainty vector for each crowd worker $w$. This uncertainty vector reveals worker $w$'s vacillation to select the local winner at each stage.

\section{Theoretical guarantees}\label{theory}
In this section, we initially present several required definitions in Section \ref{setup}. Then, we use these prerequisites to derive the key theories in Section \ref{Lseminorm-minimaxbound} and Section \ref{L2norm-minimaxbound}, which justify the robustness of DATELINE theoretically.

\subsection{Prerequisites}\label{setup}
Assume that worker $w$ annotates $N_w$ ($k$-ary) preferences with ability $\vec{\eta}_w$. The $i$-th preference ($i \in \{ 1,\cdots,N_w\}$) can be represented as a $d \times k$ matrix $E_i$, where $d$ denotes the number of all objects with the (instance-dependent) ground-truth score vector $\vec{\lambda}_{\theta}^*$, and $k$ represents the length of each preference.

Each $E_i$ positions $k$ objects to be compared, where $1$ denotes the compared object and its rank in $i$-th preference. Assume that $R_1 \cdots R_k$ as permutation matrices, and each $k \times k$ permutation matrix shift $E_i$ in a fixed direction. Therefore, we define the function $F$:

\begin{equation}
F(\vec{\lambda}_{\theta}^{*\top}E_i R_j) = \Pr(j > \{1,\cdots,j-1,j+1,\cdots,k\}),
\end{equation}
where $j \in [k]$, and $F$ denotes the probability that $j$-th object in preference $E_i$ should be chosen as the local winner according to the ground-truth score vector $\vec{\lambda}_{\theta}^*$. To simplify our analysis, we provide the first stage of DATELINE model, which can be abstracted as function $G$:

\begin{equation}
G(v(\vec{\lambda}_{\theta}^*),\eta_{w}) = \sum_{j=1}^{k}\eta_{w}^{j}F(v(\vec{\lambda}_{\theta}^*),\eta_{w}))_{|v(\vec{\lambda}_{\theta}^*) = \vec{\lambda}_{\theta}^{*\top}E_iR_j},
\end{equation}
where $\eta_{w}^{j}$ represents the probability that he/she should have selected the $j$-th object in preference $E_i$ as the local winner at the first stage. Assume that function $F$ satisfies strong log-concavity. Since function $G$ is the linear combination of function $F$, then $G$ also satisfies strong log-concavity, namely:

\begin{equation}
\begin{split}
\nabla_{\vec{\lambda}_{\theta}}^2 (-\log F(\vec{\lambda}_{\theta})) &\geq H_F,\\
\nabla_{v}^2 (-\log G(v(\vec{\lambda}_{\theta}),\vec{\eta}_{w})) &\geq H_G,
\end{split}
\end{equation}
where $H_F$ is some symmetric matrix related to function $F$ with $\lambda_2(H_F) > 0$. $H_G$ is some symmetric matrix related to function $G$ with $\lambda_2(H_G) > 0$.

\begin{definition}
Laplacian matrix L induces a semi-norm given by:
\begin{equation}
\lVert X \rVert_L = \sqrt{X^T L X}.
\end{equation}
\end{definition}

\begin{definition}\label{laplacian-matrix}
Let $L$ be an ($d \times d$) matrix that depends on the choice of the comparison topology, and $L$ represents the Laplacian of the comparison hyper-graph:
\begin{equation}
L = \frac{1}{N_w}\sum_{i=1}^{N_w}E_i(kI - 11^T)E_i^T.
\end{equation}
\end{definition}

\subsection{Minimax rates in $L$ semi-norm}\label{Lseminorm-minimaxbound}
We provide minimax rates of DATELINE in $L$ semi-norm. The proof is in Appendix~A1.
\begin{theorem}
\textbf{(Minimax rates of DATELINE in $L$ semi-norm)} Assume that $\vec{\eta}_w$ is estimated correctly, which reflects the worker ability in ground truth. (1) The estimator $\vec{\lambda}_{\theta}'$ by DATELINE has Laplacian minimax upper bound as follows:
\begin{equation}
\inf_{\vec{\lambda}_{\theta}'}\sup_{\vec{\lambda}_{\theta}^*}E[\lVert \vec{\lambda}_{\theta}' - \vec{\lambda}_{\theta}^* \rVert_L^2]\leq \frac{k^2 \sup_{v}\lVert\nabla_{v} \log G(v,\vec{\eta}_{w})\rVert_2^2}{\lambda_2(H_{G(v,\vec{\eta}_{w})})^2}\frac{(d-1)}{N_w}.
\end{equation}
(2) The estimator $\vec{\lambda}_{\theta}'$ by DATELINE has Laplacian minimax lower bound as follows:
\begin{equation}
\inf_{\vec{\lambda}_{\theta}'}\sup_{\vec{\lambda}_{\theta}^*}E[\lVert \vec{\lambda}_{\theta}' - \vec{\lambda}_{\theta}^* \rVert_L^2]\geq \frac{C(\alpha,d)\inf_{z}F(z)}{\lambda_{\max}(H_F)\sup_z \lVert \nabla F(z) \rVert_{H_F^{\dagger}}^2} \frac{d}{N_w\sup(\vec{\eta}_w)},
\end{equation}
where $C(\alpha,d) = 0.005(1-\frac{0.01d + \log2}{\log M(\alpha)})$.
\end{theorem}

\begin{remark}\label{remark-minimax}
When the worker $w$ is an expert or malicious worker, namely, $\sup(\vec{\eta}_w) \approx 1$, the lower bound is small, which means that the estimated radius centered at the optimal $\vec{\lambda}_{\theta}^*$ is small. Thus, the estimator $\vec{\lambda}_{\theta}'$ recovered by DATELINE relatively approaches the optimal $\vec{\lambda}_{\theta}^*$. However, when the worker $w$ is an amateur or spammer, namely, $\sup(\vec{\eta}_w) <$ or $\ll 1$, the lower bound is large, which means that the estimated radius centered at the optimal $\vec{\lambda}_{\theta}^*$ is large. Thus, the gap exists between the estimator $\vec{\lambda}_{\theta}'$ and the optimal $\vec{\lambda}_{\theta}^*$.
\end{remark}

\subsection{Minimax rates in $\ell_2$-norm}\label{L2norm-minimaxbound}
Inspired by the minimax rates in $L$ semi-norm, we extend the above minimax rates into $\ell_2$-norm. We can draw the similar conclusions as Remark~\ref{remark-minimax}. The proof is in Appendix~A2.
\begin{theorem}
\textbf{(Minimax rates of DATELINE in $\ell_2$-norm)} Assume that $\vec{\eta}_w$ is estimated correctly, which reflects the worker ability in ground truth. (1) The estimator $\vec{\lambda}_{\theta}'$ by DATELINE has Euclidean minimax upper bound as follows:
\begin{equation}
\inf_{\vec{\lambda}_{\theta}'}\sup_{\vec{\lambda}_{\theta}^*}E[\lVert \vec{\lambda}_{\theta}' - \vec{\lambda}_{\theta}^* \rVert_2^2]\leq \frac{k^2 \sup_{v}\lVert\nabla_{v} \log G(v,\vec{\eta}_{w})\rVert_2^2}{\lambda_2(L)\lambda_2(H_{G(v,\vec{\eta}_{w})})^2}\frac{(d-1)}{N_w}.
\end{equation}
(2) The estimator $\vec{\lambda}_{\theta}'$ by DATELINE has Euclidean minimax lower bound as follows:
\begin{equation}
\inf_{\vec{\lambda}_{\theta}'}\sup_{\vec{\lambda}_{\theta}^*}E[\lVert \vec{\lambda}_{\theta}' - \vec{\lambda}_{\theta}^* \rVert_2^2] \geq \frac{C(\alpha,d)\inf_{z}F(z)}{k(k-1)\lambda_{\max}(H_F)\sup_z \lVert \nabla F(z) \rVert_{H_F^{\dagger}}^2} \frac{d^2}{N_w\sup(\vec{\eta}_w)},
\end{equation}
where $C(\alpha,d) = 0.005(1-\frac{0.01d + \log2}{\log M(\alpha)})$.
\end{theorem}

\section{Conclusions}\label{conclusion}
This paper introduces a new setting in preference aggregation called instance-dependent noisy preferences. This practical setting not only considers the feature information of ranking objects, but also considers the dynamic size of preferences (object comparison). Based on this new setting, we propose \textbf{D}eep pl\textbf{A}cke\textbf{T}t-luce mod\textbf{EL} w\textbf{I}th u\textbf{N}certainty m\textbf{E}asurements (DATELINE). Namely, we leverage deep neural networks mapping each instance into its ranking score of Plackett-Luce model, and design a weighted Plackett-Luce model to overcome the uncertainty in $k$-ary noisy preferences. Besides, we provide theoretical guarantees for DATELINE to justify its robustness. In future, we will collect instance-dependent noisy preferences setting in the real world, and conduct experiments on this setting by DATELINE. This will justify the robustness of DATELINE in practice.

\appendix

\subsection*{A0: Required Lemmas}
\begin{lemma}\label{laplacian-trace}
The Laplacian matrix meets the trace constraints, namely, \mbox{$\nullspace(L) = 1$}, the eigenvalue $\lambda_2(L) > 0$, and
\begin{equation}
\tr(L) = k(k-1).
\end{equation}
\end{lemma}

\begin{lemma}
For any $j \in [k]$, $i \in [N_w]$ and any vector $v \in \mathbb{R}^k$, we have,
\begin{equation}
\frac{\lambda_2(H)}{k}v^{\top}(kI - 11^{\top})v \leq v^{\top} R_j H R_j^{\top} v \leq \frac{\lambda_{\max}(H)}{k}v^{\top}(kI - 11^{\top})v,
\end{equation}
where $H$ is a symmetric matrix with $\lambda_2(H) > 0$. Note that, $H$ can be set as $H_F$ or $H_G$.
\end{lemma}

\begin{lemma}\label{strong-covexity}
\textbf{(Upper bound for M-estimators)} Consider the estimator $\widehat{\Omega}$, where $\widehat{\Omega} \in \arg\min_{\Omega}l(\Omega)$. If $l$ is a differentiable function satisfying the $\kappa$-strong convexity at optimal $\Omega^*$, then we have:
\begin{equation}
\lVert \widehat{\Omega} - \Omega^*\rVert_L \leq \frac{1}{\kappa}\lVert \nabla_{\Omega^*} l(\Omega^*) \rVert_{L^{\dagger}}.
\end{equation}
\end{lemma}

\begin{lemma}\label{Binary-GVB}
\textbf{(Binary Gilbert-Varshamov bound)} For any $\alpha \in (0,\frac{1}{4})$, if there is a subset $\mathcal{V}$ of the $d$-dimensional hypercube \mbox{$H_d = \{0,1\}^d$}, where $\mathcal{V} = \{z^1,\cdots,z^{M(\alpha)}\}$ and $M(\alpha) = \exp\{\frac{d}{2}(\log 2 + 2\alpha\log 2\alpha + (1-2\alpha)\log(1-2\alpha))\}$, then we have
\begin{equation}
\begin{split}
\alpha d \leq \lVert z^\phi - z^\varphi \rVert_2^2 &\leq d;\\
\langle e_1, z^\phi \rangle &= 0,
\end{split}
\end{equation}
where all $\phi \neq \varphi$ $\in [M(\alpha)]$, and $e_1$ denotes the first canonical basis vector.
\end{lemma}

\begin{lemma}\label{Generalized-GVB}
\textbf{(Generalized Gilbert-Varshamov bound)} For any $\alpha \in (0,\frac{1}{4})$, assume that there is a subset $\mathcal{V}$ of the $d$-dimensional hypercube, where $\mathcal{V} = \{\Omega^1,\cdots,\Omega^{M(\alpha)}\}$ and $M(\alpha) = \exp\{\frac{d}{2}(\log 2 + 2\alpha\log 2\alpha + (1-2\alpha)\log(1-2\alpha))\}$. Let $L$ come from Definition \ref{laplacian-matrix}, where $L$ can be decomposed as $U^T \Lambda U$, $U$ is an orthonormal matrix, and $\Lambda$ is a diagnal matrix. If $\Omega^\phi = \frac{\delta}{\sqrt{d}}U^T \Lambda^{\dagger} z^\phi$ for $\phi \in [M(\alpha)]$ and $z^\phi \in \{0,1\}^d$, then we have
\begin{equation}
\alpha \delta^2 \leq \lVert \Omega^{\phi} - \Omega^{\varphi} \rVert_L^2 \leq \delta^2,
\end{equation}
where all $\phi \neq \varphi$ $\in [M(\alpha)]$.
\end{lemma}

\begin{lemma}\label{Fano-minimax}
\textbf{(Generalized Fano minimax bound)} For any $\alpha \in (0,\frac{1}{4})$, suppose that we can construct a $\delta$-packing in $\rho$-semimetric with cardinality $M(\alpha) = \exp\{\frac{d}{2}(\log 2 + 2\alpha\log 2\alpha + (1-2\alpha)\log(1-2\alpha))\}$. Namely, we have a packing set $\mathcal{V} = \{\Omega^1,\cdots,\Omega^{M(\alpha)}\}$, and each pair from this set meets \mbox{$\alpha \delta^2 \leq \lVert \Omega^{\phi} - \Omega^{\varphi} \rVert_L^2 \leq \delta^2$} where $\phi \neq \varphi$ $\in [M(\alpha)]$ (generalized Gilbert-Varshamov bound). Then the generalized Fano minimax risk between the estimator $\widehat{\Omega}$ and the optimal $\Omega^*$ has lower bound as follows:

\begin{equation}
\inf_{\widehat{\Omega}}\sup_{\Omega^*}E[\rho(\widehat{\Omega}, \Omega^*)^2] \geq \frac{\delta^2}{2}(1-\frac{\bar{D}_{KL}(\mathbb{P}_{\Omega^\phi} || \mathbb{P}_{\Omega^\varphi}) + \log2}{\log M(\alpha)}),
\end{equation}
where $\bar{D}_{KL}(\mathbb{P}_{\Omega^\phi} || \mathbb{P}_{\Omega^\varphi}) = \sum_{i=1}^{N_w}\sum_{l=1}^{k}\eta_w^l F(\Omega^{\phi^T} E_iR_l)\log\frac{F(\Omega^{\phi^T} E_iR_l)}{F(\Omega^{\varphi^T} E_iR_l)}$ is weighted KL divergence between distributions $\mathbb{P}_{\Omega^\phi}$ and $\mathbb{P}_{\Omega^\varphi}$, considering the worker quality $\vec{\eta}_w$.
\end{lemma}

\subsection*{A1: Proof of Minimax Rates in $L$ Semi-norm}\label{Minimax-Semi-norm}
\paragraph{Upper Bound}\label{UB-L-SemiNorm}
We prove this upper bound by using Lemma~\ref{strong-covexity}. The log likelihood in this paper can be written as:
\begin{equation}
l(\vec{\lambda}_{\theta}) = -\frac{1}{N_w}\sum_{i=1}^{N_w}\log G(v(\vec{\lambda}_{\theta}),\vec{\eta}_{w})_{|v(\vec{\lambda}_{\theta}) = \vec{\lambda}_{\theta}^{\top} E_iR_j}.
\end{equation}
Therefore, the $\nabla_{\vec{\lambda}_{\theta}} l(\vec{\lambda}_{\theta})$ is:
\begin{equation}
\begin{split}
\nabla_{\vec{\lambda}_{\theta}} l(\vec{\lambda}_{\theta}) &= -\frac{1}{N_w}\sum_{i=1}^{N_w} \nabla_{\vec{\lambda}_{\theta}} \log G(v(\vec{\lambda}_{\theta}),\vec{\eta}_{w})\\
&= -\frac{1}{N_w}\sum_{i=1}^{N_w} \nabla_{v} \log G(v(\vec{\lambda}_{\theta}),\vec{\eta}_{w})E_iR_j.
\end{split}
\end{equation}
Also, the $\nabla_{\vec{\lambda}_{\theta}}^2 l(\vec{\lambda}_{\theta})$ is:
\begin{equation}
\nabla_{\vec{\lambda}_{\theta}}^2 l(\vec{\lambda}_{\theta}) = -\frac{1}{N_w}\sum_{i=1}^{N_w} E_iR_j \nabla_{v}^2\log G(v(\vec{\lambda}_{\theta}),\vec{\eta}_{w})R_j^{\top} E_i^{\top}.
\end{equation}
For any vector $z \in \mathbb{R}^d$, we have
\begin{equation}
\begin{split}
z^T\nabla_{\vec{\lambda}_{\theta}}^2 l(\vec{\lambda}_{\theta})z &= -\frac{1}{N_w}\sum_{i=1}^{N_w}z^{\top} E_iR_j \nabla_{v}^2\log G(v(\vec{\lambda}_{\theta}),\vec{\eta}_{w})R_j^{\top} E_i^{\top} z\\
&\geq \frac{1}{N_w}\sum_{i=1}^{N_w}z^{\top} E_iR_j H_G R_j^{\top} E_i^{\top} z\\
&\geq \frac{1}{N_w}\sum_{i=1}^{N_w}z^{\top}  \frac{\lambda_2(H_G)}{k}E_i^{\top}(kI - 11^{\top})E_i z\\
&= \frac{\lambda_2(H_G)}{k}z^{\top}\frac{1}{N_w}\sum_{i=1}^{N_w}E_i^{\top}(kI - 11^{\top})E_i z\\
&= \frac{\lambda_2(H_G)}{k} \lVert z \rVert_L^2.
\end{split}
\end{equation}
Therefore, $l$ is verified as the $\kappa$-strong convexity, where $\kappa = \frac{\lambda_2(H_G)}{k}$. According to Lemma~\ref{strong-covexity}, we have:

\begin{equation}
\lVert \vec{\lambda}_{\theta}' - \vec{\lambda}_{\theta}^*\rVert_L^2 \leq \frac{k^2}{\lambda_2(H_G)^2}\lVert \nabla_{\vec{\lambda}_{\theta}^*} l(\vec{\lambda}_{\theta}^*) \rVert_{L^{\dagger}}^2.
\end{equation}
Therefore, the key for above equation is to upperbound $\lVert \nabla_{\vec{\lambda}_{\theta}^*} l(\vec{\lambda}_{\theta}^*) \rVert_{L^{\dagger}}^2$. Now, we rewrite the gradient of log likelihood at $\vec{\lambda}_{\theta}^*$ as:

\begin{equation}
\nabla_{\vec{\lambda}_{\theta}^*} l(\vec{\lambda}_{\theta}^*) = -\frac{1}{N_w}\sum_{i=1}^{N_w} E_iV_i,
\end{equation}
where $V_i = \nabla_{v} \log G(v(\vec{\lambda}_{\theta}^*),\eta_{w})R_j$. If we define $M = I - \frac{1}{k}11^{\top}$, then $L = \frac{k}{N_w}\sum_{i=1}^{N_w}E_i M E_i^{\top}$. Since $M$ is a symmetric matrix, then pseudo-inverse $M^{\dagger} = M$. Define $\widetilde{V}_i = (M^{\dagger})^{\frac{1}{2}}V_i$ for each $i \in [n]$. Consider the shift invariance property, the function $g(t) = G(x+t1,\eta_w) - G(x,\eta_w)$ is constant, thus we have,

\begin{equation}
g'(t) = \langle \nabla_{v} G(v,\vec{\eta}_w), 1 \rangle = 0.
\end{equation}
Therefore, we have

\begin{equation}
\langle \nabla_{v} \log G(v,\vec{\eta}_w), 1 \rangle = \langle \frac{\nabla_{v} G(v,\vec{\eta}_w)}{G(v,\vec{\eta}_w)}, 1 \rangle = 0.
\end{equation}
Here, we rewrite the log likelihood as:

\begin{equation}
\nabla_{\vec{\lambda}_{\theta}^*} l(\vec{\lambda}_{\theta}^*) = -\frac{1}{N_w}\sum_{i=1}^{N_w} E_i(M)^{\frac{1}{2}}\widetilde{V}_i.
\end{equation}
Therefore, we have
\begin{equation}
\nabla_{\vec{\lambda}_{\theta}^*} l(\vec{\lambda}_{\theta}^*) L^{\dagger} \nabla_{\vec{\lambda}_{\theta}^*} l(\vec{\lambda}_{\theta}^*) = \frac{1}{N_w^2}\sum_{i=1}^{N_w}\sum_{i'=1}^{N_w} \widetilde{V}_i^{\top}(M)^{\frac{1}{2}}E_i^{\top} L^{\dagger}E_{i'}(M)^{\frac{1}{2}}\widetilde{V}_{i'}.
\end{equation}
Before delving into the expectation of $\nabla_{\vec{\lambda}_{\theta}^*} l(\vec{\lambda}_{\theta}^*) L^{\dagger} \nabla_{\vec{\lambda}_{\theta}^*} l(\vec{\lambda}_{\theta}^*)$, we calculate the expectation of $\widetilde{V}_i$ first:

\begin{equation}
\begin{split}
\mathbb{E}[\widetilde{V}_i] &= \mathbb{E}[(M^{\dagger})^{\frac{1}{2}}V_i]\\
&= \mathbb{E}[(M^{\dagger})^{\frac{1}{2}} R_j \nabla_{v} \log G(v(\vec{\lambda}_{\theta}^*),\vec{\eta}_{w})]\\
&= (M^{\dagger})^{\frac{1}{2}} \frac{1}{k}\sum_{j=1}^{k}R_j \nabla_{v} \log G(v(\vec{\lambda}_{\theta}^*),\vec{\eta}_{w})_{|v(\vec{\lambda}_{\theta}^*) = \vec{\lambda}_{\theta}^{*\top}E_iR_j}\\
&= (M^{\dagger})^{\frac{1}{2}} \frac{1}{k}\sum_{j=1}^{k} R_j \langle \nabla_{v} \log G(v(\vec{\lambda}_{\theta}^*),\vec{\eta}_{w}), 1\rangle = 0.
\end{split}
\end{equation}
%
Therefore, we now calculate $\mathbb{E}[\nabla_{\vec{\lambda}_{\theta}^*} l(\vec{\lambda}_{\theta}^*) L^{\dagger} \nabla_{\vec{\lambda}_{\theta}^*} l(\vec{\lambda}_{\theta}^*)]$ as follows:
\begin{equation}
\begin{split}
& \mathbb{E}[\nabla_{\vec{\lambda}_{\theta}^*} l(\vec{\lambda}_{\theta}^*) L^{\dagger} \nabla_{\vec{\lambda}_{\theta}^*} l(\vec{\lambda}_{\theta}^*)]\\
& = \mathbb{E}[\frac{1}{N_w^2}\sum_{i=1}^{N_w}\sum_{i'=1}^{N_w} \widetilde{V}_i^{\top}(M)^{\frac{1}{2}}E_i^{\top} L^{\dagger}E_{i'}(M)^{\frac{1}{2}}\widetilde{V}_{i'}]\\
&= \frac{1}{N_w^2}\mathbb{E}[\sum_{i=1,i'=i}^{N_w}\widetilde{V}_i^{\top}(M)^{\frac{1}{2}}E_i^{\top} L^{\dagger}E_{i'}(M)^{\frac{1}{2}}\widetilde{V}_{i'} + \sum_{i=1,i'\neq i}^{N_w}\widetilde{V}_i^{\top}(M)^{\frac{1}{2}}E_i^{\top} L^{\dagger}E_{i'}(M)^{\frac{1}{2}}\widetilde{V}_{i'}]\\
&= \frac{1}{N_w^2}\mathbb{E}[\sum_{i=1,i'=i}^{N_w}\widetilde{V}_i^{\top}(M)^{\frac{1}{2}}E_i^{\top} L^{\dagger}E_{i'}(M)^{\frac{1}{2}}\widetilde{V}_{i'}],
\end{split}
\end{equation}
where the last equality corrects due to $\mathbb{E}[\widetilde{V}_i]\mathbb{E}[\widetilde{V}_{i'}] = 0$ when $i\neq i'$. To sum up, we have
\begin{equation}
\begin{split}
\mathbb{E}[\nabla_{\vec{\lambda}_{\theta}^*} l(\vec{\lambda}_{\theta}^*) L^{\dagger} \nabla_{\vec{\lambda}_{\theta}^*} l(\vec{\lambda}_{\theta}^*)] &= \frac{1}{N_w^2}\mathbb{E}[\sum_{i=1}^{N_w}\widetilde{V}_i^{\top}(M)^{\frac{1}{2}}E_i^{\top} L^{\dagger}E_{i}(M)^{\frac{1}{2}}\widetilde{V}_{i}]\\
&\leq \frac{1}{N_w}\mathbb{E}[\sup_{i\in [n]} (\widetilde{V}_i^{\top}\widetilde{V}_{i})] \frac{1}{N_w}\mathbb{E}[\sum_{i=1}^{N_w}(M)^{\frac{1}{2}}E_i^{\top} L^{\dagger}E_{i}(M)^{\frac{1}{2}}]\\
&\leq \frac{1}{N_w}\mathbb{E}[\sup_{i\in [n]} (\widetilde{V}_i^{\top}\widetilde{V}_{i})] \frac{1}{N_w} tr(\sum_{i=1}^{N_w}(M)^{\frac{1}{2}}E_i^{\top} L^{\dagger}E_{i}(M)^{\frac{1}{2}}).
\end{split}
\end{equation}
Here, we calculate $\widetilde{V}_i^{\top}\widetilde{V}_{i}$,

\begin{equation}
\begin{split}
\widetilde{V}_i^{\top}\widetilde{V}_{i} &= [(M^{\dagger})^{\frac{1}{2}} R_j \nabla_{v} \log G(v(\vec{\lambda}_{\theta}^*),\vec{\eta}_{w})]^{\top} (M^{\dagger})^{\frac{1}{2}} R_j \nabla_{v} \log G(v(\vec{\lambda}_{\theta}^*),\vec{\eta}_{w})\\
&= \nabla_{v} \log G(v(\vec{\lambda}_{\theta}^*),\vec{\eta}_{w}) R_j^{\top} M^{\dagger} R_j \nabla_{v} \log G(v(\vec{\lambda}_{\theta}^*),\vec{\eta}_{w})\\
&= \lVert\nabla_{v} \log G(v(\vec{\lambda}_{\theta}^*),\vec{\eta}_{w})\rVert_M^2,
\end{split}
\end{equation}
where $R_j^{\top} M^{\dagger} R_j = M^{\dagger} = M$. Since $L = \frac{k}{N_w}\sum_{i=1}^{N_w}E_i M E_i^{\top}$, we have

\begin{equation}
\frac{1}{N_w} \tr(\sum_{i=1}^{N_w}(M)^{\frac{1}{2}}E_i^{\top} L^{\dagger}E_{i}(M)^{\frac{1}{2}}) = \frac{d-1}{k}.
\end{equation}
Therefore, we have the upper bound for $\mathbb{E}[\nabla_{\vec{\lambda}_{\theta}^*} l(\vec{\lambda}_{\theta}^*) L^{\dagger} \nabla_{\vec{\lambda}_{\theta}^*} l(\vec{\lambda}_{\theta}^*)]$, namely:

\begin{equation}
\begin{split}
\mathbb{E}[\nabla_{\vec{\lambda}_{\theta}^*} l(\vec{\lambda}_{\theta}^*) L^{\dagger} \nabla_{\vec{\lambda}_{\theta}^*} l(\vec{\lambda}_{\theta}^*)]
&\leq \frac{1}{N_w}\mathbb{E}[\sup_{i\in [n]} (\widetilde{V}_i^{\top}\widetilde{V}_{i})] \frac{1}{N_w} \tr(\sum_{i=1}^{N_w}(M)^{\frac{1}{2}}E_i^{\top} L^{\dagger}E_{i}(M)^{\frac{1}{2}})\\
&\leq \frac{d-1}{k} \frac{1}{N_w}\mathbb{E}[\sup_{i\in [n]}\lVert\nabla_{v} \log G(v(\vec{\lambda}_{\theta}^*),\vec{\eta}_{w})\rVert_M^2]\\
&\leq \frac{d-1}{k} \frac{1}{N_w}k\sup_{v}\lVert\nabla_{v} \log G(v(\vec{\lambda}_{\theta}^*),\vec{\eta}_{w})\rVert_2^2\\
&= \frac{d-1}{N_w} \sup_{v}\lVert\nabla_{v} \log G(v(\vec{\lambda}_{\theta}^*),\vec{\eta}_{w})\rVert_2^2.
\end{split}
\end{equation}
To sum up, we have:

\begin{equation}
\begin{split}
\mathbb{E}\lVert \vec{\lambda}_{\theta}' - \vec{\lambda}_{\theta}^*\rVert_L^2 &\leq \frac{k^2}{\lambda_2(H_G)^2}E[\lVert \nabla_{\vec{\lambda}_{\theta}^*} l(\vec{\lambda}_{\theta}^*) \rVert_{L^{\dagger}}^2\\
&\leq \frac{k^2}{\lambda_2(H_G)^2}\frac{d-1}{N_w} \sup_{v}\lVert\nabla_{v} \log G(v(\vec{\lambda}_{\theta}^*),\vec{\eta}_{w})\rVert_2^2\\
&\leq \frac{k^2 (d-1) \sup_{v}\lVert\nabla_{v} \log G(v(\vec{\lambda}_{\theta}^*),\vec{\eta}_{w})\rVert_2^2}{\lambda_2(H_G)^2 N_w}.
\end{split}
\end{equation}

\paragraph{Lower Bound}\label{LB-L-SemiNorm}
For any pair of quality score vectors $\vec{\lambda}_{\theta}^{\phi}$ and $\vec{\lambda}_{\theta}^{\varphi}$, the weighted KL divergence, considering the worker quality $\vec{\eta}_w$, between the distributions $\mathbb{P}_{\vec{\lambda}_{\theta}^\phi}$ and $\mathbb{P}_{\vec{\lambda}_{\theta}^\varphi}$ is:

\begin{equation}
\begin{split}
\bar{D}_{KL}(\mathbb{P}_{\vec{\lambda}_{\theta}^\phi} || \mathbb{P}_{\vec{\lambda}_{\theta}^\varphi}) &= \sum_{i=1}^{N_w}\sum_{l=1}^{k}\eta_w^l F(\vec{\lambda}_{\theta}^{\phi\top} E_iR_l)\log\frac{F(\vec{\lambda}_{\theta}^{\phi\top} E_iR_l)}{F(\vec{\lambda}_{\theta}^{\varphi\top} E_iR_l)}\\
&\leq \sum_{i=1}^{N_w}\sum_{l=1}^{k}\eta_w^l F(\vec{\lambda}_{\theta}^{\phi\top} E_iR_l)\log\frac{F(\vec{\lambda}_{\theta}^{\phi\top} E_iR_l)}{F(\vec{\lambda}_{\theta}^{\varphi\top} E_iR_l)}\\
&\leq \sum_{i=1}^{N_w}\sum_{l=1}^{k}\eta_w^l F(\vec{\lambda}_{\theta}^{\phi\top} E_iR_l)(\frac{F(\vec{\lambda}_{\theta}^{\phi\top} E_iR_l)}{F(\vec{\lambda}_{\theta}^{\varphi\top} E_iR_l)} - 1),
\end{split}
\end{equation}
where the last equation is due to $\log x \leq x-1$. Since the fact that $\sum_{l=1}^{m}F(\vec{\lambda}_{\theta}^{\phi\top} E_iR_l) = \sum_{l=1}^{m}F(\vec{\lambda}_{\theta}^{\varphi\top} E_iR_l) = 1$, then we have

\begin{equation}
\begin{split}
\bar{D}_{KL}(\mathbb{P}_{\vec{\lambda}_{\theta}^\phi} || \mathbb{P}_{\vec{\lambda}_{\theta}^\varphi})
&\leq \sum_{i=1}^{N_w}\sum_{l=1}^{k}\eta_w^l F(\vec{\lambda}_{\theta}^{\phi\top} E_iR_l)(\frac{F(\vec{\lambda}_{\theta}^{\phi\top} E_iR_l)}{F(\vec{\lambda}_{\theta}^{\varphi\top} E_iR_l)} - 1)\\
&\leq \sum_{i=1}^{N_w}\sum_{l=1}^{k}\sup(\vec{\eta}_w)(\frac{F(\vec{\lambda}_{\theta}^{\phi\top} E_iR_l)^2}{F(\vec{\lambda}_{\theta}^{\varphi\top} E_iR_l)} - F(\vec{\lambda}_{\theta}^{\phi\top} E_iR_l))\\
&= \sum_{i=1}^{N_w}\sum_{l=1}^{k}\sup(\vec{\eta}_w)(\frac{F(\vec{\lambda}_{\theta}^{\phi\top} E_iR_l)^2}{F(\vec{\lambda}_{\theta}^{\varphi\top} E_iR_l)} - 2F(\vec{\lambda}_{\theta}^{\phi\top} E_iR_l) + F(\vec{\lambda}_{\theta}^{\varphi\top} E_iR_l))\\
&= \sup(\vec{\eta}_w)\sum_{i=1}^{N_w}\sum_{l=1}^{k} \frac{(F(\vec{\lambda}_{\theta}^{\phi\top} E_iR_l) - F(\vec{\lambda}_{\theta}^{\varphi\top} E_iR_l))^2}{F(\vec{\lambda}_{\theta}^{\varphi\top} E_iR_l)}\\
&\leq \frac{\sup(\vec{\eta}_w)}{\inf_{z}F(z)}\sum_{i=1}^{N_w}\sum_{l=1}^{k}(F(\vec{\lambda}_{\theta}^{\phi\top} E_iR_l) - F(\vec{\lambda}_{\theta}^{\varphi\top} E_iR_l))^2\\
&\leq \frac{\sup(\vec{\eta}_w)}{\inf_{z}F(z)}\sum_{i=1}^{N_w}\sum_{l=1}^{k}(\langle \nabla F(z_{il}), \vec{\lambda}_{\theta}^{\phi\top} E_iR_l - \vec{\lambda}_{\theta}^{\varphi\top} E_iR_l\rangle)^2,
\end{split}
\end{equation}
where the last equation corrects due to $F$ is assumed to be strongly log-concave. Therefore, we have

\begin{equation}\small
\begin{split}
\bar{D}_{KL}(\mathbb{P}_{\vec{\lambda}_{\theta}^\phi} || \mathbb{P}_{\vec{\lambda}_{\theta}^\varphi})
&\leq \frac{\sup(\vec{\eta}_w)}{\inf_{z}F(z)}\sum_{i=1}^{N_w}\sum_{l=1}^{k}(\langle \nabla F(z_{il}), \vec{\lambda}_{\theta}^{\phi\top} E_iR_l - \vec{\lambda}_{\theta}^{\varphi\top} E_iR_l\rangle)^2\\
&\leq \frac{\sup(\vec{\eta}_w)\sup_z \lVert \nabla F(z) \rVert_{H_F^{\dagger}}^2}{\inf_{z}F(z)}\sum_{i=1}^{N_w}\sum_{l=1}^{k} \lVert \vec{\lambda}_{\theta}^{\phi\top} E_iR_l - \vec{\lambda}_{\theta}^{\varphi\top} E_iR_l\rVert_{H_F}^2\\
&\leq \frac{\sup(\vec{\eta}_w)\sup_z \lVert \nabla F(z) \rVert_{H_F^{\dagger}}^2}{\inf_{z}F(z)}(\vec{\lambda}_{\theta}^{\phi} - \vec{\lambda}_{\theta}^{\varphi})^{\top} \{\sum_{i=1}^{N_w}\sum_{l=1}^{k} E_i^{\top} R_l H_F R_l^{\top} E_i\}(\vec{\lambda}_{\theta}^{\phi} - \vec{\lambda}_{\theta}^{\varphi})\\
&\leq \frac{\sup(\vec{\eta}_w)\sup_z \lVert \nabla F(z) \rVert_{H_F^{\dagger}}^2}{\inf_{z}F(z)}(\vec{\lambda}_{\theta}^{\phi} - \vec{\lambda}_{\theta}^{\varphi})^{\top} \{\sum_{i=1}^{N_w}\sum_{l=1}^{k} \frac{\lambda_{\max}(H_F)}{k}E_i^{\top}(kI - 11^{\top})E_i\}(\vec{\lambda}_{\theta}^{\phi} - \vec{\lambda}_{\theta}^{\varphi})\\
&= \frac{\sup(\vec{\eta}_w)\sup_z \lVert \nabla F(z) \rVert_{H_F^{\dagger}}^2}{\inf_{z}F(z)}\frac{\lambda_{\max}(H_F)}{k}(\vec{\lambda}_{\theta}^{\phi} - \vec{\lambda}_{\theta}^{\varphi})^{\top} \{\sum_{i=1}^{N_w}\sum_{l=1}^{k} E_i^{\top}(kI - 11^{\top})E_i\}(\vec{\lambda}_{\theta}^{\phi} - \vec{\lambda}_{\theta}^{\varphi})\\
&= \frac{N_w\lambda_{\max}(H_F)\sup(\vec{\eta}_w)\sup_z \lVert \nabla F(z) \rVert_{H_F^{\dagger}}^2}{\inf_{z}F(z)}(\vec{\lambda}_{\theta}^{\phi} - \vec{\lambda}_{\theta}^{\varphi})^{\top} \{\frac{k}{N_w}\sum_{i=1}^{N_w} E_i^{\top}(I - \frac{1}{k}11^{\top})E_i\}(\vec{\lambda}_{\theta}^{\phi} - \vec{\lambda}_{\theta}^{\varphi})\\
&= \frac{N_w\lambda_{\max}(H_F)\sup(\eta_w)\sup_z \lVert \nabla F(z) \rVert_{H_F^{\dagger}}^2}{\inf_{z}F(z)}\lVert\vec{\lambda}_{\theta}^{\phi} - \vec{\lambda}_{\theta}^{\varphi}\rVert_L^2.
\end{split}
\end{equation}
Based on Lemma~\ref{Generalized-GVB}, we have

\begin{equation}
\begin{split}
\bar{D}_{KL}(\mathbb{P}_{\vec{\lambda}_{\theta}^\phi} || \mathbb{P}_{\vec{\lambda}_{\theta}^\varphi})
&\leq \frac{N_w\lambda_{\max}(H_F)\sup(\vec{\eta}_w)\sup_z \lVert \nabla F(z) \rVert_{H_F^{\dagger}}^2}{\inf_{z}F(z)}\lVert\vec{\lambda}_{\theta}^{\phi} - \vec{\lambda}_{\theta}^{\varphi}\rVert_L^2\\
&\leq \frac{N_w\lambda_{\max}(H_F)\sup(\vec{\eta}_w)\sup_z \lVert \nabla F(z) \rVert_{H_F^{\dagger}}^2}{\inf_{z}F(z)}\delta^2
\end{split}
\end{equation}
Therefore, when $\delta^2 = \frac{0.01d \inf_{z}F(z)}{N_w\lambda_{\max}(H_F)\sup(\vec{\eta}_w)\sup_z \lVert \nabla F(z) \rVert_{H_F^{\dagger}}^2}$, we have $\bar{D}_{KL}(\mathbb{P}_{\vec{\lambda}_{\theta}^\phi} || \mathbb{P}_{\vec{\lambda}_{\theta}^\varphi}) \leq 0.01d$. Based on Lemma~\ref{Fano-minimax}, when we choose $\rho = \lVert \cdot \rVert_L$, we have the lower bound:

\begin{equation}
\inf_{\vec{\lambda}_{\theta}'}\sup_{\vec{\lambda}_{\theta}^*}E[\lVert \vec{\lambda}_{\theta}' - \vec{\lambda}_{\theta}^* \rVert_L^2] \geq \frac{0.005d \inf_{z}F(z)}{N_w\lambda_{\max}(H_F)\sup(\vec{\eta}_w)\sup_z \lVert \nabla F(z) \rVert_{H_F^{\dagger}}^2}(1-\frac{0.01d + \log2}{\log M(\alpha)}).
\end{equation}

\subsection*{A2: Proof of Minimax Rates in $\ell_2$-norm}

\paragraph{Upper Bound}\label{UB-L2-Norm}
Note that $(\vec{\lambda}_{\theta}' - \vec{\lambda}_{\theta}^*)$ $\bot$ nullspace($L$), then we have $\lVert \vec{\lambda}_{\theta}' - \vec{\lambda}_{\theta}^* \rVert_L^2 \geq \lambda_2(L)\lVert \vec{\lambda}_{\theta}' - \vec{\lambda}_{\theta}^* \rVert_2^2$. Therefore, $l_2$-norm minimax upper bound of DATELINE is:

\begin{equation}
\begin{split}
\inf_{\vec{\lambda}_{\theta}'}\sup_{\vec{\lambda}_{\theta}^*}E[\lVert \vec{\lambda}_{\theta}' - \vec{\lambda}_{\theta}^* \rVert_2^2] &\leq \frac{1}{\lambda_2(L)}\inf_{\vec{\lambda}_{\theta}'}\sup_{\vec{\lambda}_{\theta}^*}E[\lVert \vec{\lambda}_{\theta}' - \vec{\lambda}_{\theta}^* \rVert_L^2]\\
&\leq \frac{k^2 \sup_{v}\lVert\nabla_{v} \log G(v,\vec{\eta}_{w})\rVert_2^2}{\lambda_2(L)\lambda_2(H_{G(v,\vec{\eta}_{w})})^2}\frac{(d-1)}{N_w}.
\end{split}
\end{equation}

\paragraph{Lower Bound}\label{LB-L2-Norm}
Based on Lemma~\ref{Binary-GVB}, we reconstruct a packing set $\{\vec{\lambda}_{\theta}^1,\cdots,\vec{\lambda}_{\theta}^{M(\alpha)} \}$, where $\vec{\lambda}_{\theta}^\phi = \frac{\delta}{\sqrt{d}}U^T R z^\phi$ for $\phi \in [M(\alpha)]$, $z^\phi \in \{0,1\}^d$ and $R$ is a permutation matrix. Then, we have:

\begin{equation}
\lVert \vec{\lambda}_{\theta}^{\phi} - \vec{\lambda}_{\theta}^{\varphi} \rVert_L^2 = \frac{\delta^2}{d}\lVert z^\phi - z^\varphi\rVert_{\Lambda} = \frac{\delta^2}{d}\sum_{i=2}^{d}\lambda_i(L) \leq \frac{\delta^2}{d}\tr(L),
\end{equation}
where $\phi, \varphi \in [M(\alpha)]$ and $L = U^{\top} \Lambda U$. Based on Lemma~\ref{laplacian-trace}, we have

\begin{equation}
\begin{split}
\bar{D}_{KL}(\mathbb{P}_{\vec{\lambda}_{\theta}^\phi} || \mathbb{P}_{\vec{\lambda}_{\theta}^\varphi})
&\leq \frac{N_w\lambda_{\max}(H_F)\sup(\vec{\eta}_w)\sup_z \lVert \nabla F(z) \rVert_{H_F^{\dagger}}^2}{\inf_{z}F(z)}\lVert\vec{\lambda}_{\theta}^{\phi} - \vec{\lambda}_{\theta}^{\varphi}\rVert_L^2\\
&\leq \frac{N_w\lambda_{\max}(H_F)\sup(\vec{\eta}_w)\sup_z \lVert \nabla F(z) \rVert_{H_F^{\dagger}}^2}{\inf_{z}F(z)}\frac{\delta^2}{d}\tr(L)\\
&= \frac{N_w\lambda_{\max}(H_F)\sup(\vec{\eta}_w)\sup_z \lVert \nabla F(z) \rVert_{H_F^{\dagger}}^2}{\inf_{z}F(z)}\frac{\delta^2}{d}k(k-1),
\end{split}
\end{equation}
Therefore, when $\delta^2 = \frac{0.01d^2\inf_{z}F(z)}{k(k-1)N_w\lambda_{\max}(H_F)\sup(\vec{\eta}_w)\sup_z \lVert \nabla F(z) \rVert_{H_F^{\dagger}}^2}$, we have \mbox{$\bar{D}_{KL}(\mathbb{P}_{\vec{\lambda}_{\theta}^\phi} || \mathbb{P}_{\vec{\lambda}_{\theta}^\varphi}) \leq 0.01d$}. Based on Lemma~\ref{Fano-minimax}, when we choose $\rho = \lVert \cdot \rVert_2$, we have the lower bound:

\begin{equation}
\inf_{\vec{\lambda}_{\theta}'}\sup_{\vec{\lambda}_{\theta}^*}E[\lVert \vec{\lambda}_{\theta}' - \vec{\lambda}_{\theta}^* \rVert_2^2] \geq \frac{0.005d^2\inf_{z}F(z)}{k(k-1)N_w\lambda_{\max}(H_F)\sup(\vec{\eta}_w)\sup_z \lVert \nabla F(z) \rVert_{H_F^{\dagger}}^2}(1-\frac{0.01d + \log2}{\log M(\alpha)}).
\end{equation}


\begin{thebibliography}{4}
\bibitem{barbera1978preference} Salvador, B. and Hugo, S.: Preference aggregation with randomized social orderings. Journal of Economic Theory. (1978)
\bibitem{bottero2018choquet} Bottero, M. and Ferretti, V. and Figueira, J. and Greco, S. and Roy, B.: On the Choquet multiple criteria preference aggregation model: Theoretical and practical insights from a real-world application. European Journal of Operational Research. (2018)
\bibitem{li2018hybrid} Li, J. and Mantiuk, R. and Wang, J. and Ling, S. and Le Callet, P.: Hybrid-MST: A hybrid active sampling strategy for pairwise preference aggregation. In: NeurIPS. (2018)
\bibitem{raman2014methods} Raman, K. and Joachims, T.: Methods for ordinal peer grading. In: KDD. (2014)
\bibitem{bartels1996uninformed} Bartels, L。: Uninformed votes: Information effects in presidential elections. American Journal of Political Science. (1996)
\bibitem{dwork2001rank} Dwork, C. and Kumar, R. and Naor, M. and Sivakumar, D.: Rank aggregation methods for the web. In: WWW. (2001)
\bibitem{volkovs2012flexible} Volkovs, M. and Zemel, R.: A flexible generative model for preference aggregation. In: WWW. (2012)
\bibitem{chen2013pairwise} Chen, X. and Bennett, P. and Collins-Thompson, K. and Horvitz, E.: Pairwise ranking aggregation in a crowdsourced setting. In: WSDM. (2013)
\bibitem{khetan2016data} Khetan, A. and Oh, S.: Data-driven rank breaking for efficient rank aggregation. Journal of Machine Learning Research. (2016)
\bibitem{soufiani2014computing} Soufiani, H. and Parkes, D. and Xia, L.: Computing parametric ranking models via rank-breaking. In: ICML. (2014)
\bibitem{han2018robust} Han, B. and Pan, Y. and Tsang, I.: Robust Plackett--Luce model for k-ary crowdsourced preferences. Machine Learning. (2018)
\bibitem{guiver2009bayesian} Guiver, J. and Snelson, E.: Bayesian inference for Plackett-Luce ranking models. In: ICML. (2009)
\bibitem{maystre2015fast} Maystre, L. and Grossglauser, M.: Fast and accurate inference of Plackett-Luce models. In: NeurIPS. (2015)
\bibitem{yan2013fast} Yan, W. and Wu, Q. and Liang, J. and Chen, Y. and Fu, X.: How fast are the leaked facial expressions: The duration of micro-expressions. Journal of Nonverbal Behavior. (2013)
\bibitem{cheng2010label} Cheng, W. and H{\"u}llermeier, E. and Dembczynski, K.: Label ranking methods based on the Plackett-Luce model. In: ICML. (2010)
\bibitem{tkachenko2016plackett} Tkachenko, M. and Lauw, H.: Plackett-Luce regression mixture model for heterogeneous rankings. In: CIKM. (2016)

\end{thebibliography}
\end{document}